\documentclass[11pt]{article}
\usepackage{array,ifthen,amsmath,amssymb,amsthm,latexsym,color,epsfig,bm,verbatim}
\usepackage{algorithm2e}

\def\eps{\varepsilon}

\def\<{\langle}
\def\>{\rangle}
\def\({\left(} 
\def\){\right)} 

\newtheorem{thm}{Theorem}
\newtheorem{que}[thm]{Question}
\newtheorem{pro}[thm]{Proposition}
\newtheorem{obs}[thm]{Observation}
\newtheorem{exc}{Exercise}
\newtheorem{nte}{Note}
\newtheorem{rem}[thm]{Remark}
\newtheorem{lem}[thm]{Lemma}
\newtheorem{cor}[thm]{Corollary}

\newtheorem{con}{Conjecture}
\newtheorem{exm}[thm]{Example}
\newtheorem{dfn}[thm]{Definition}

\newenvironment{prf}[1]{\noindent{\bf{Proof #1\\}}}{$\hfill\blacksquare$\nopagebreak[4]\vskip 0.3cm}

\def\argmin{\mathrm{argmin}}

\def\BS{\begin{footnotesize}}
\def\ES{\end{footnotesize}}
\def\BCON{\begin{con}}
\def\ECON{\end{con}}
\def\BOBS{\begin{obs}}
\def\EOBS{\end{obs}}

\def\BPRO{\begin{pro}}
\def\EPRO{\end{pro}}
\def\BIDE{\begin{ide}}
\def\EIDE{\end{ide}}
\def\BTHM{\begin{thm}}
\def\ETHM{\end{thm}}
\def\BDEF{\begin{dfn}\rm}
\def\EDEF{\end{dfn}}
\newcommand\BPRF[1][:]{\begin{prf}{#1}}
\def\EPRF{\end{prf}}
\def\BLEM{\begin{lem}}
\def\ELEM{\end{lem}}
\def\BEX{\begin{exm}\rm}
\def\EEX{\end{exm}}
\def\BEXC{\begin{exc}\rm}
\def\EEXC{\end{exc}}
\def\BCOR{\begin{cor}}
\def\ECOR{\end{cor}}
\def\BQUE{\begin{que}}
\def\EQUE{\end{que}}
\newcommand\BSOL[1][:]{\begin{sol}{#1}}
\def\ESOL{\end{sol}}
\def\BNTE{\begin{nte}}
\def\ENTE{\end{nte}}

\def\BIT{\begin{itemize}}
\def\EIT{\end{itemize}}
\def\BREM{\begin{rem}\rm}
\def\EREM{\end{rem}}
\def\BC{\begin{center}}
\def\EC{\end{center}}
\def\BEQ{\begin{equation}}
\def\EEQ{\end{equation}}

\date{}

\title{Density estimation in linear time}

\author{
Satyaki Mahalanabis\thanks{
Department of Computer Science, University of Rochester,
Rochester, NY 14627.  Email: \{smahalan,stefanko\}@cs.rochester.edu}
\and
Daniel \v{S}tefankovi\v{c}$^*$
}

\begin{document}

\maketitle

\begin{abstract}
We consider the problem of choosing a density estimate from a set
of distributions~${\cal F}$, minimizing the $L_1$-distance to an
unknown distribution (\cite{DG01}). Devroye and Lugosi~\cite{DG01}
analyze two algorithms for the problem: Scheff\'e tournament winner
and minimum distance estimate. The Scheff\'e tournament estimate
requires fewer computations than the minimum distance estimate,
but has strictly weaker guarantees than the latter.

We focus on the computational aspect of density estimation.
We present two algorithms, both with the same guarantee as the
minimum distance estimate. The first one, a modification of
the minimum distance estimate, uses the same number (quadratic in
$|{\cal F}|$) of computations as the Scheff\'e tournament.
The second one, called ``efficient minimum loss-weight estimate,''
uses only a linear number of computations, assuming that ${\cal
F}$ is preprocessed.

We also give examples showing that the guarantees of the algorithms
cannot be improved and explore randomized algorithms for density estimation.
\end{abstract}

\section{Introduction}
\label{sec:intro}

We study the following density estimation problem considered
in~\cite{DG96,DG01,DL02}. There is an unknown distribution $g$ and
we are given $n$ (not necessarily independent) samples which
define empirical distribution $h$. Given a finite class $\cal{F}$
of distributions, our objective is to output $f \in \cal{F}$ such
that the error $\|f - g \|_1$ is minimized. The use of the
$L_1$-norm is well justified by it has many useful properties, for
example, scale invariance and the fact that approximate
identification of a distribution in the $L_1$-norm gives an
estimate for the probability of every event.

The following two parameters influence the error of a possible
estimate: the distance of $g$ from ${\cal F}$ and the empirical
error. The first parameter is required since we have no control
over $\cal{F}$, and hence we cannot select a distribution which
is better than the ``optimal'' distribution in $\cal{F}$, that is,
the one closest to $g$ in $L_1$-norm. It is not obvious how to
define the second parameter---the error of $h$ with respect to $g$.
We follow the definition of~\cite{DG01}, which is inspired
by~\cite{Y85} (see Section~\ref{sec:intro:def} for a precise definition).

Devroye and Lugosi~\cite{DG01} analyze two algorithms in this
setting: Scheff\'e tournament winner and minimum distance
estimate. The minimum distance estimate, defined by
Yatracos~\cite{Y85}, is a special case of the minimum distance
principle, formalized by Wolfowitz in~\cite{Wolf57}. The minimum
distance estimate is a helpful tool, for example, it was used by~\cite{DG96,DG97}
to obtain estimates for the smoothing factor for kernel density estimates
and also by~\cite{DL02} for hypothesis testing.

The Scheff\'e tournament winner algorithm requires fewer
computations than the minimum distance estimate, but it has strictly
weaker guarantees (in terms of the two parameters mentioned above)
than the latter. Our main contribution are two procedures for
selecting an estimate from $\cal{F}$, both of which have the same
guarantees as the minimum distance estimate, but are
computationally more efficient. The first has a quadratic (in
$|{\cal F}|$) cost, matching the cost of the Scheff\'e tournament
winner algorithm. The second one is even faster, using {\em linearly}
many (in $|{\cal F}|$) computations (after preprocessing ${\cal F}$).

Now we outline the rest of the paper.
In Section~\ref{sec:intro:def} we give the required definitions and
introduce the notion of a test-function (a variant of Scheff\'e set).
Then, in Section~\ref{previo}, we restate the previous density estimation
algorithms (Scheff\'e tournament winner and the minimum distance estimate)
using test-functions. Next, in Section~\ref{yyyyy}, we present our
algorithms. The first one is a modification of the minimum-distance estimate
with improved (quadratic in $|\cal F|$) computational cost. The second
one, which we call ``efficient minimum loss-weight estimate,'' has only
{\em linear} computational cost after preprocessing ${\cal F}$.
In Section~\ref{srandom} we explore randomized density estimation
algorithms. In the final Section~\ref{selow}, we give examples showing
tightness of the theorems stated in the previous sections.

Throughout this paper we focus on the case when ${\cal F}$ is finite, in
order to compare the computational costs of our estimates to
previous ones. However our results generalize in a straightforward
way to infinite classes as well if we ignore computational
complexity.

\subsection{Definitions and Notations}\label{sec:intro:def}

Throughout the paper $g$ will be the unknown distribution and $h$ will be the empirical distribution.
Let $\cal{F}$ be a set of distributions. We will assume that ${\cal F}$ is finite (the results generalize
straightforwardly to infinite sets of distributions). Let ${\rm d}_1(g, {\cal F})$ be the $L_1$-distance
of $g$ from ${\cal F}$, that is, $\min_{f \in {\cal F}} \| f - g \|_1$.

Given two functions $f_i, f_j$ on $\Omega$ (in this context, distributions) we define a
\emph{test-function} $T_{ij}:\Omega\rightarrow\{-1,0,1\}$
to be the function $T_{ij}(x)={\rm sgn}(f_i(x) - f_j(x))$. Note that $T_{ij} = -T_{j\,i}$. We also define
${\cal T}_{\cal F}$ to be the set of all test-functions for ${\cal F}$, that is,
\begin{equation*}\label{testf}
{\cal T}_{\cal F}=\{\, T_{ij}\,\,|\,\, f_i,f_j\in {\cal F}\,\}.
\end{equation*}
Let $\cdot$ be the inner product for the functions on $\Omega$. Note that
\begin{equation*}\label{l1test}
(f_i-f_j)\cdot T_{ij} = \|f_i-f_j\|_1.
\end{equation*}
We use the inner product of the empirical distribution $h$ with the test-functions to choose an estimate,
which is a distribution from ${\cal F}$.

In this paper we only consider algorithms which make their decisions purely on inner products of
the test-functions with $h$ and members of ${\cal F}$. It is reasonable to assume that the
computation of the inner product will take significant time. Hence we measure the {\em computational cost}
of an algorithm is by the number of inner products used.

We say that $f_i$ {\em wins} against $f_j$ if
\begin{equation}\label{lo}
(f_i-h)\cdot T_{ij} < (f_j-h) \cdot T_{j\,i}.
\end{equation}
Note that either $f_i$ wins against $f_j$, or $f_j$ wins against $f_i$, or there is a draw
(that is, there is equality in \eqref{lo}).

The algorithms choose an estimate $f\in{\cal F}$ using the empirical distribution $h$.
The $L_1$-distance of the estimates from the unknown distribution $g$ will depend on the
following measure of distance between the empirical and the unknown distribution:
\begin{equation}\label{emp}
\Delta:=\max\limits_{T\in{\cal T}_{\cal F}} (g-h)\cdot T.
\end{equation}

Now we discuss how test-functions can be viewed as a reformulation of Scheff\'e sets,
defined by Devroye and Lugosi~\cite{DG01} (inspired by~\cite{S47} and implicit in~\cite{Y85}),
as follows.
The Scheff\'e set of distributions $f_i,f_j$ is
$$A_{ij}=\{x\,\,;\,\, f_i(x)>f_j(x)\}.$$
Devroye and Lugosi say that $f_i$ wins against $f_j$ if
\begin{equation}\label{lug}
\left| \int_{A_{ij}} f_i - h(A_{ij}) \right| <
\left| \int_{A_{ij}} f_j - h(A_{ij}) \right|.
\end{equation}
The advantage of using Scheff\'e sets is that for a concrete set ${\cal F}$ of distributions
one can immediately use the theory of Vapnik-Chervonenkis dimension~\cite{VC71}
for the family of Scheff\'e sets of ${\cal F}$ (this family is called the {\em Yatracos class}
of ${\cal F}$), to obtain a bound on the empirical error.

If $h,f_i,f_j$ are distributions then the condition \eqref{lo} is {\em equivalent} to \eqref{lug}
(to see this recall that $T_{ij}=-T_{j\,i}$, and add
$(f_i-h)\cdot {\mathbf 1}=(h-f_j)\cdot {\mathbf 1}$ to \eqref{lo}, where ${\mathbf 1}$ is the
vector of all ones). Thus, in our algorithms the test-functions
can be replaced by Scheff\'e sets and VC dimension arguments can be applied.

We chose to use test-functions for two reasons: first, they allow us to give
succinct proofs of our theorems (especially Theorem~\ref{t4}),
and second, they immediately extend to the case when the members of
${\cal F}$ are not distributions (cf, e.\,g., Exercise~6.2, in~\cite{DG01}).

\BREM
Note that our value of $\Delta$, defined in terms of ${\cal
T}_{\cal F}$, is at most twice the $\Delta$ used in \cite{DG01},
which is defined in terms of Scheff\'e sets.
\EREM

\subsection{Previous Estimates}\label{previo}

In this section we restate the two algorithms for density estimation from
Chapter 6 of \cite{DG01}) using test-functions. The first algorithm requires
less computation but has worse guarantees than the second algorithm.

\begin{center}
\fbox{\begin{minipage}{12cm}
{\bf Algorithm 1} - {\sc Scheff\'e tournament winner}.\\
Output $f\in {\cal F}$ with the most wins (tie broken arbitrarily).
\end{minipage}}\\
\vskip 0.2cm
\end{center}

\BTHM[\cite{DG01}, Theorem~6.2]\label{t1}
Let $f_1\in {\cal F}$ be the distribution output by Algorithm~1.
Then $$\|f_1-g\|_1\leq 9\,{\rm d}_1(g,{\cal F})+8\Delta.$$
The number of inner products used by Algorithm~1 is $\Theta(|{\cal F}|^2)$.
\ETHM

\begin{center}
\fbox{\begin{minipage}{12cm}
{\bf Algorithm 2} - {\sc Minimum distance estimate}.\\
Output $f\in {\cal F}$ that minimizes
\begin{equation}\label{a2}
\max\big\{\,| (f-h)\cdot T_{ij} |\,\,;\,\,f_i,f_j\in {\cal F}\,\big\}.
\end{equation}
\end{minipage}}\\
\vskip 0.2cm
\end{center}

\BTHM[\cite{DG01}, Theorem~6.3]\label{t2}
Let $f_1$ be the distribution output by Algorithm~2.
Then $$\|f_1-g\|_1\leq 3\,{\rm d}_1(g,{\cal F})+2\Delta.$$
The number of inner products used by Algorithm~2 is $\Theta(|{\cal F}|^3)$.
\ETHM

Let us point out that Theorems~6.2~and~6.3 in \cite{DG01} require that each $f \in \cal{F}$
is a distribution, that is, $\int f = 1$. Since we use test-functions in the algorithms instead
of Scheff\'e set based comparisons, the assumption $\int f=1$ is not actually needed in
the proofs of Theorems~6.2~and~6.3 (we skip the proof), and is not used in the proofs
of Theorems~\ref{t3},~\ref{t4}.

\section{Our estimators}\label{yyyyy}

\subsection{A variant of the minimum distance estimate}\label{varia}

The following modified minimum distance estimate uses only
$O(|{\cal F}|^2)$ computations as compared to $O(|{\cal F}|^3)$
computations used by Algorithm~2 (equation~\eqref{a3} takes
minimum of $O(|{\cal F}|)$ terms, whereas equation~\eqref{a2}
takes minimum of $O(|{\cal F}|^2)$ terms), but as we show in
Theorem ~\ref{t3}, it gives us the same guarantee as the minimum
distance estimate.

\begin{center}
\fbox{\begin{minipage}{12cm}
{\bf Algorithm 3} - {\sc Modified minimum distance estimate}.\\
Output $f_i\in {\cal F}$ that minimizes
\begin{equation}\label{a3}
\max\big\{\,|(f_i-h)\cdot T_{ij}|\,\,;\,\,f_j\in {\cal F}\,\big\}.
\end{equation}
\end{minipage}}\\
\vskip 0.2cm
\end{center}

\BTHM\label{t3}
Let $f_1\in {\cal F}$ be the distribution output by Algorithm~3.
Then $$\|f_1-g\|_1\leq 3\,{\rm d}_1(g,{\cal F})+2\Delta.$$
The number of inner products used by Algorithm~3 is $\Theta(|{\cal F}|^2)$.
\ETHM

\BPRF
Let $f_1\in {\cal F}$ be the function output by Algorithm 3. Let $f_{2} = \argmin_{f \in \mathcal{F}} \|f - g\|_1$.
By the triangle inequality we have
\begin{equation}\label{eaaa}
\|f_1 - g\|_1 \leq \|f_1 - f_2\|_1 + \|f_2 - g\|_1.
\end{equation}
We bound $\|f_1 - f_2\|_1$ as follows:
\begin{equation*}
\begin{split}
\|f_1 - f_2\|_1 = (f_1 - f_2) \cdot T_{12} \leq |(f_1 - h) \cdot
T_{12}| + |(f_2 - h) \cdot T_{12}| \\ \leq |(f_1 - h) \cdot
T_{12}| + \max_{f_j\in \mathcal{F}} |(f_2 - h) \cdot T_{2,j}|,
\end{split}
\end{equation*}
where in the last inequality we used the fact that $T_{12}=-T_{21}$.

By the criteria of selecting $f_1$ we have
$|(f_1 - h) \cdot T_{12}| \leq \max_{f_j\in \mathcal{F}} |(f_2 - h) \cdot T_{2,j}|$
(since otherwise $f_2$ would be selected). Hence
\begin{equation*}
\begin{split}
\|f_1 - f_2\|_1 \leq 2 \max_{f_j\in \mathcal{F}} |(f_2 - h) \cdot T_{2,j}|
\leq 2 \max_{f_j\in \mathcal{F}} |(f_2 - g) \cdot  T_{2,j}| + 2 \max_{f_j\in \mathcal{F}} |(g - h) \cdot T_{2,j}| \\ \leq
2\|(f_2 - g)\|_1 + 2 \max_{T \in \mathcal{T_{F}}} |(g - h) \cdot T| = 2 \|f_2 - g\|_1 + 2 \Delta.
\end{split}
\end{equation*}
Combining the last inequality with \eqref{eaaa} we obtain
\[ \|f_1 - g\|_1 \leq 3\|f_2 - g\|_1 + 2 \Delta. \]
\EPRF

\BREM\label{rem}
Note that one can modify the Lemma to only require that $g$ and $h$ be ``close'' with respect to the
test functions for the ``best'' function in the class, that is, only $|(g-h)\cdot T_{2,j}|$ need
to be small (where $f_2$ is $\argmin_{f \in \mathcal{F}} \|f - g\|_1$).
\EREM

One can ask whether the observation in Remark~\ref{rem} can lead
to improved density estimation algorithms for concrete sets of
distributions. The bounds on $\Delta$ (which is given by
\eqref{emp}) are often based on the VC-dimension of the Yatracos
class of ${\cal F}$. Recall that the Yatracos class $Y$ is the set of
$A_{ij}=\{x\,;\,f_i(x)>f_j(x)\}$ for all $f_i,f_j\in{\cal F}$.
Remark~\ref{rem} implies that instead of the Yatracos class it is
enough to consider the set $Y_i=\{A_{ij}\,;\,f_j\in{\cal F}\}$ for
$f_i\in{\cal F}$. Is it possible that the VC-dimension of each set
$Y_i$ is smaller the VC-dimension of the Yatracos class $Y$?
The following (artificial) example shows that this can, indeed,
be the case. Let $\Omega=\{0,\dots,n\}$. For each $(n+1)$-bit binary
string $a_0,a_1,\dots,a_n$, let us consider the distribution
$$
P(k)= \frac{1}{4n}(1 + (1/2-a_0)(1/2-a_k)) 2^{-\sum_{j=1}^n a_j 2^j},
$$
for $k\in\{1,\dots,n\}$ (with $P(0)$ chosen to make $P$ into a distribution).
For this family of $2^{n+1}$ distributions the VC-dimension of the Yatracos class is $n$, whereas
each $Y_i$ has VC-dimension $1$ (since a pair of distributions $f_i,f_j$ has
a non-trivial set $A_{ij}$ if and only if their binary strings differ only in
the first bit).

\subsection{An even more efficient estimator - minimum loss-weight}\label{seffice}

In this section we present an estimator which, after preprocessing ${\cal F}$,
uses only $O(|{\cal F}|)$ inner products to obtain a density estimate. The
guarantees of the estimate are the same as for Algorithms 2 and 3.

The algorithm uses the following quantity to choose the estimate:
\begin{equation*}\label{a4}
\mbox{\rm loss-weight}(f)=\max\big\{\,\|f-f'\|_1\,\,;\,\, f\ \mbox{does not win against}\ f'\in{\cal F}\,\big\}.
\end{equation*}

Intuitively a good estimate should have small loss-weight (ideally
the loss-weight of the estimate would be $-\infty=\max\{\}$, that is, the
estimate would not lose at all). Thus the following algorithm
would be a natural candidate for a good density estimator (and,
indeed, it has a guarantee matching Algorithms 2 and 3), but,
unfortunately, we do not know how to implement it using $O(|{\cal
F}|)$ inner products.
\begin{center}
\fbox{\begin{minipage}{12cm}
{\bf Algorithm 4a} - {\sc Minimum loss-weight estimate}.\\
Output $f\in {\cal F}$ that minimizes $\mbox{\rm loss-weight}(f)$.
\end{minipage}}\\
\vskip 0.2cm
\end{center}
The next algorithm, seems less natural than algorithm~4a, but its condition can be implemented using only
$O(|{\cal F}|)$ inner products.
\begin{center}
\fbox{\begin{minipage}{12cm}
{\bf Algorithm 4b} - {\sc Efficient minimum loss-weight estimate}.\\
Output $f\in {\cal F}$ such that for every $f'$ to which $f$ loses we have
\begin{equation}\label{a5}
\|f-f'\|_1\leq\mbox{\rm loss-weight}(f').
\end{equation}
\end{minipage}}\\
\vskip 0.2cm
\end{center}

Before we delve into the proof of \eqref{et4} let us see how
Algorithm~4b can be made to use $|{\cal F}|-1$ inner products. We
preprocess ${\cal F}$ by computing $L_1$-distances between all
pairs of distributions in ${\cal F}$ and store the distances in an
list sorted in decreasing order. When the algorithm is presented with the empirical
distribution $h$, all it needs to do is perform comparison between select pairs of distributions.
The advantage is that we preprocess ${\cal F}$ only once and, for each new empirical distribution
we only compute inner products necessary for the comparisons.

We will compute the estimate as follows.

\incmargin{1em} \restylealgo{boxed}\linesnumbered
\begin{algorithm}[H]\label{a:1}\small
\dontprintsemicolon \SetKwInOut{Input}{input} \SetKwInOut{Output}{output}
\SetKwInOut{Assume}{assume\phantom{a}}
\Input{
family of distributions ${\cal F}$, list $L$ of all pairs $\{f_i,f_j\}$ sorted in decreasing order by
$\|f_i-f_j\|_1$, oracle for computing inner products $h\cdot T_{ij}$.}
\Output{$f\in{\cal F}$ such that: $(\forall f')$
$f$ loses to $f'$ $\implies$ $\|f-f'\|_1\leq\mbox{\rm loss-weight}(f')$.}
\BlankLine
$S\leftarrow {\cal F}$\;
\Repeat{$|S|=1$}{
pick the first edge $\{f_i,f_j\}$ in $L$\;
{\bf if} $f_i$ loses to $f_j$ {\bf then} $f'\leftarrow f_i$ {\bf else} $f'\leftarrow f_j$ {\bf fi}\;
 remove $f'$ from $S$\;
 remove pairs containing $f'$ from $L$}
output the distribution in $S$\;
\label{algo_disjdecomp}
\end{algorithm}
\vspace{-0.45cm}
\begin{center}
{\bf Algorithm 4b - using $O(|{\cal F}|)$ inner products.}
\end{center}
\decmargin{1em}

Note that while Algorithm 4b~uses only $O(|{\cal F}|)$ inner products its running time
is actually $\Theta(|{\cal F}|^2)$, since it traverses a list of length $\Theta(|{\cal F}|^2)$.
If we are willing to spend exponential time for the preprocessing then we can build the
complete decision tree corresponding to Algorithm~4b and obtain a linear-time density
selection procedure. Is it possible to achieve linear running time
using only polynomial-time preprocessing?

\BQUE[Tournament Revelation Problem] We are given a weighted
undirected complete graph on $n$ vertices. Assume that the
edge-weights are distinct. We preprocess the weighted graph and
then play the following game with an adversary until only one
vertex remains: we report the edge with the largest weight and the
adversary chooses one of the endpoints of the edge and removes it
from the graph (together with all the adjacent edges).

Our goal is to make the computational cost during the game linear-time (in $n$)
in the worst-case (over the adversary's moves). Is it possible to achieve this
goal with polynomial-time preprocessing?
\EQUE

We now show that estimate $f$ output by algorithm~4b
satisfies \eqref{a5} for every $f'$ against which $f$ loses. We
show, using induction, that the following invariant is always
satisfied on line $2$. For any $f\in S$ and any $f'\in {\cal
F}\setminus S$ we have that if $f$ loses to $f'$ then
$\|f-f'\|_1\leq\mbox{\rm loss-weight}(f')$. Initially, ${\cal
F}\setminus S$ is empty and the invariant is trivially true. For
the inductive step, let $f'$ be the distribution most recently
removed from $S$. To prove the induction step we only need to show
that for every $f\in S$ we have that if $f$ loses to $f'$ then
$\|f-f'\|_1\leq\mbox{\rm loss-weight}(f')$. Let $W$ be the
$L_1$-distance between two distributions in $S\cup\{f'\}$. Then
loss-weight$(f')\geq W$ (since $f'$ lost), and $\|f-f'\|_1\leq W$
(by the definition of $W$).
\BTHM\label{t4}
Let $f_1\in {\cal F}$ be the distribution output by Algorithm~4a (or Algorithm~4b).
Then
\begin{equation}\label{et4}
\|f_1-g\|_1\leq 3\,{\rm d}_1(g,{\cal F})+2\Delta.
\end{equation}
Assume that we are given $L_1$-distances between every pair in ${\cal F}$.
The number of inner products used by Algorithm~4b is $\Theta(|{\cal F}|)$.
\ETHM

\BPRF[of Theorem~\ref{t4}:]
Let $f_4=g$. Let $f_2$ be the function $f\in {\cal F}$ minimizing $\|g-f\|_1$. We can
reformulate our goal \eqref{et4} as follows:
\begin{equation}\label{goal}
(f_1-f_4)\cdot T_{14} \leq 2\Delta+ 3(f_2-f_4)\cdot T_{24}.
\end{equation}
Let $f_3\in{\cal F}$ be the function $f'\in {\cal F}$ such that $f_2$ loses against $f_3$ and
$\|f_{2}-f'\|_1$ is maximal.  Note that
$f_1,f_2,f_3\in{\cal F}$, but $f_4$ does need to be in ${\cal F}$.

We know that $f_2$ loses against $f_3$, that is, we have (see \eqref{lo})
\begin{equation}\label{e2}
2h\cdot T_{23} \leq f_2\cdot T_{23} + f_3\cdot T_{23},
\end{equation}
and, since $f_1$ minimized the maximum loss, we also have
\begin{equation}\label{e3}
(f_1-f_2)\cdot T_{12}\leq (f_2-f_3)\cdot T_{23}.
\end{equation}
By \eqref{emp} we have
\begin{equation}\label{e4}
2(f_4-h)\cdot T_{23}\leq 2\Delta.
\end{equation}
Adding \eqref{e2}, \eqref{e3}, and \eqref{e4} we obtain
\begin{equation}\label{e5}
2(f_2-f_4)\cdot T_{23}+ (f_2-f_1)\cdot T_{12} + 2\Delta\geq 0.
\end{equation}

Note that for any $i,j,k,\ell$ we have:
\begin{equation}\label{eeep}
(f_i-f_j)\cdot (T_{ij}-T_{k\ell})\geq 0,
\end{equation}
since if $f_i(x)>f_j(x)$ then $T_{ij}-T_{k\ell}\geq 0$, if $f_i(x)<f_j(x)$ then $T_{ij}-T_{k\ell}\leq 0$,
and if $f_i(x)=f_j(x)$ then the contribution of that $x$ is zero. By applying \eqref{eeep} four times
we obtain
\begin{equation}\label{e6}
(f_2-f_4)\cdot (3 T_{24}-2T_{23}-T_{14}) + (f_1-f_2)\cdot (T_{12}-T_{14})\geq 0.
\end{equation}

Finally, adding \eqref{e5} and \eqref{e6} yields \eqref{goal}.
\EPRF

\BREM
Note that Remark~\ref{rem} also applies to Algorithms~4a and 4b, since \eqref{e4} is the only inequality
in which $\Delta$ is used.
\EREM

\BREM
If the condition~\eqref{a5} of Algorithm~4b is relaxed to
\begin{equation}\label{a5nn}
\|f-f'\|_1\leq C\cdot \mbox{\rm loss-weight}(f'),
\end{equation}
for some $C\geq 1$, one can prove an analogue of Theorem~\ref{t4}
with \eqref{et4} replaced by
\begin{equation}\label{et4nn}
\|f_1-g\|_1\leq (1+2C)\,{\rm d}_1(g,{\cal F})+2C\Delta.
\end{equation}
\EREM

\section{Randomized algorithm and mixtures}\label{srandom}

In this section we explore the following question: can
constant $3$ be improved if we allow randomized algorithms?
Let $f$ be the output of a randomized algorithm ($f$ is a random
variable with values in ${\cal F}$). We would like to bound the
expected error ${\mathrm E}\big[\| f - g \|_1\big]$.

If instead of randomization we consider algorithms which output
mixtures of distributions in ${\cal F}$ we obtain a related
problem. Indeed, let $\alpha$ be the distribution on ${\cal F}$
produced by a randomized algorithm, and let
$r=\sum_{s\in{\cal F}} \alpha_s s$ be the corresponding mixture.
Then, by triangle inequality, we have
$$
\|r-g\|_1\leq {\mathrm E}\big[ \|f-g\|_1\big].
$$
Hence the model in which the output is allowed to be a mixture of
distributions in ${\cal F}$ is ``easier'' than the model
in which the density selection algorithm is randomized.

We consider here only the special case in which ${\cal F}$ has only two
distributions $f_1, f_2$, and give an randomized algorithm
with a better guarantee than is possible for deterministic
algorithms. Later, in Section~\ref{selow}, we give a matching lower
bound in the mixture model.

To simplify the exposition we will, without loss of generality, assume that $\|f_1-f_2\|_1>0$.
Thus for any $h$ we have $(f_1-h)\cdot T_{12} + (h-f_2)\cdot T_{12} = \|f_1-f_2\|_1 > 0$.
\begin{center}
\fbox{\begin{minipage}{12cm}
{\bf Algorithm 5 - {\sc Randomized estimate}.}\\
Let
\begin{equation*}\label{alteq1}
r=\frac{|(f_{1}-h) \cdot T_{12}|}{|(f_{2}-h) \cdot T_{12}|}.
\end{equation*}
With probability $1/(r+1)$ output $f_1$, otherwise output $f_2$.
\end{minipage}}\\
\vskip 0.2cm
\end{center}
(By convention, if $|(f_{2}-h) \cdot T_{12}|=0$ then we take $r=\infty$ and output $f_2$ with probability~$1$).

\BTHM\label{mixthm2}
Let ${\cal F}=\{f_1,f_2\}$. Let $f\in {\cal F}$ be the distribution output by Algorithm~5.
Then $${\mathrm E}\Big[\|f-g\|_1\Big]\leq 2\,{\rm d}_1(g,{\cal F})+\Delta.$$
\ETHM

\BPRF
Without loss of generality assume that $f_{2} = \argmin_{f \in \mathcal{F}} \|f - g\|_1$. First
we bound the error of $f_1$ and later use it to bound the error of $f$. We have, by triangle inequality,
$$\|f_{1} - g\|_1 \leq \|f_{1} - f_{2}\|_1 + \|f_{2}-g \|_1.$$
We can bound $\|f_{1} - f_{2}\|_1$ as follows
\begin{equation*}
\begin{split}
\|f_{1} - f_{2}\|_1 = (f_1 - f_2) \cdot T_{12} \leq |(f_{1} - h) \cdot T_{12}| + |(f_{2} - h) \cdot T_{12}|\\
= (r + 1) |(f_{2} - h) \cdot T_{12}| \leq (r + 1) |(f_{2} - g) \cdot T_{12}| + (r + 1)|(g - h) \cdot T_{12}|.
\end{split}
\end{equation*}
Thus
\begin{equation}
\label{alteq2}
\|f_{1} - g\|_1 \leq (r + 2)\|f_{2} - g\|_1 + (r + 1) \Delta.
\end{equation}
Hence
\[  {\mathrm E}\Big[\|f - g\|_1\Big] = \frac{1}{r + 1} \|f_{1} - g\|_1 + \frac{r}{r+1} \|f_{2} - g\|_1
\leq 2 \|f_{2} - g\|_1 + \Delta, \]
where in the last inequality we used~\eqref{alteq2}.
\EPRF

\section{Lower bound examples}\label{selow}

In this section we construct an example showing that deterministic distribution selection
algorithms based on test-functions cannot improve on the constant $3$,
that is, Theorems~\ref{t1}, \ref{t2}, \ref{t3}, \ref{t4} are tight. For algorithms that output
mixtures (and hence randomized algorithms) the example yields a lower bound of $2$, matching the
constant in Theorem~\ref{mixthm2}.

\BLEM\label{lww}
For every $\eps'>0$ there exist distributions $f_1,f_2$, and $g=h$ such that
\begin{equation*}
\|f_1-g\|_1 \geq (3-\eps')\|f_2-g\|_1,
\end{equation*}
and $f_1\cdot T_{12}=-f_2\cdot T_{12}$ and $h\cdot T_{12}=0$.
\ELEM

Before we prove Lemma~\ref{lww} let us see how it is applied.
Consider the behavior of the algorithm on empirical distribution $h$ for
${\cal F}=\{f_1,f_2\}$ and ${\cal F'}=\{f'_1,f'_2\}$, where $f'_1=f_2$ and $f'_2=f_1$. Note that
$T'_{12}=T_{21}=-T_{12}$ and hence
$$f'_1\cdot T'_{12}=-f'_2\cdot T'_{12}=f_1\cdot T_{12} = - f_2\cdot T_{12}.$$
Moreover, we have $h\cdot T_{12} = h\cdot T'_{12} = 0$. Note that all the
test-functions have the same value for ${\cal F}$ and ${\cal F}'$.
Hence a test-function based algorithm either outputs $f_1$ and $f'_1$, or it outputs
$f_2$ and $f'_2=f_1$. In both cases it outputs $f_1$ for one of the inputs and
hence we obtain the following consequence.

\BCOR
For any $\eps>0$ and any deterministic test-function based algorithm there exist an input
${\cal F}$ and $h=g$ such that the output $f_1$ of the algorithm satisfies
$\|f_1-g\|_1\geq (3-\eps){\mathrm d}_1(g,{\cal F})$.
\ECOR

\BPRF[of Lemma~\ref{lww}:]
Consider the following probability space consisting of of $4$ atomic events $A_1, A_2, A_3, A_4$:
\begin{center}
\begin{tabular}{c|c|c|c|c|}
    & $A_1$         & $A_2$         & $A_3$     & $A_4$ \\
\hline
$f_1$   & $0$           & $1/4 + \eps$  & $1/2$ & $1/4 - \eps$ \\
$f_2$   & $1/2+\eps$    & $1/4 - \eps$  & $0$       & $1/4$ \\
$g = h$ & $1/2$     & $1/2$     & $0$       & $0$       \\
\hline
$T_{12}$ & $-1$     & $1$     & $1$       & $-1$       \\
\end{tabular}
\end{center}
Note that we have $f_1 \cdot T_{12} = - f_2 \cdot T_{12} = \frac{1}{2} +
2\eps$, and $\| f_1 - g \|_1 = \frac{3}{2} - 2\eps, \| f_2 - g
\|_1 = \frac{1}{2} + \eps$. The ratio
$\| f_1 - g \|_1 / \| f_2 - g
\|_1$ gets arbitrarily close to $3$ as $\eps$ goes to zero.
\EPRF

Consider $f_1$ and $f_2$ from the proof of Lemma~\ref{lww}. Let
$f=\alpha f_1 + (1-\alpha)f_2$ where $\alpha\geq 1/2$. For
$0<\eps<1/4$ we have $\|f-g\|_1 = 1/2+\alpha
- 2\eps \alpha\geq 1-2\eps$. By symmetry, for one of
${\cal F}=\{f_1,f_2\}$ and ${\cal F'}=\{f'_1,f'_2\}$ (with
$f'_1=f_2$ and $f'_2=f_1$), the algorithm outputs
$\alpha f_1+(1-\alpha) f_2$ with $\alpha\geq 1/2$,
and hence we obtain the following.

\BCOR
For any $\eps>0$ and any deterministic test-function based algorithm which outputs
a mixture there exist an input ${\cal F}$ and $h=g$ such that the output $f$ of the algorithm satisfies
$\|f-g\|_1\geq (2-\eps){\mathrm d}_1(g,{\cal F})$.
\ECOR

Thus for two distributions the correct constant is $2$ for randomized algorithms using test-functions.
For larger families of distributions we do not know what the value of the
constant is (we only know that it is from the interval $[2,3]$).

\BQUE
What is the correct constant for
deterministic test-function based algorithm which output a mixture? What is the correct constant for
randomized test-function based algorithms?
\EQUE

Next we construct an example showing that $9$ is the right constant for Algorithm~1.

\BLEM
For every $\eps'>0$ there exist probability distributions $f_1,f_2,f_3=f'_3$ and $g$
such that
$$\|f_1-g\|_1 \geq (9-\eps')\|f_2-g\|_1,$$
yet the Algorithm~1, for  ${\cal F}=\{f_1,f_2,f_3,f'_3\}$, even when given the true
distribution (that is, $h=g$) outputs $f_1$.
\ELEM

\BPRF
Consider the following probability space with $6$ events $A_1,\dots,A_6$ and $f_1,f_2$ and $g$ with the
probabilities given by the following table:
\begin{center}
\begin{tabular}{c|c|c|c|c|c|c|}
    & $A_1$ & $A_2$ & $A_3$ & $A_4$ & $A_5$ & $A_6$ \\
  \hline
  $g=h$ & $2/3-21\eps$ & $1/9-2\eps$ & $9\eps$       & 0             & $2/9+14\eps$ & 0 \\
  $f_1$ & 0            & $18\eps$    & $2/3-12\eps$  & $2/9-13\eps$  & $9\eps$      & $1/9-2\eps$ \\
  $f_2$ & $2/3-30\eps$ & 0           & 0             & 0             & $2/9+14\eps$ & $1/9+16\eps$ \\
  $f_3$ & $2/3-21\eps$ & $9\eps$     & $9\eps$       & $2/9-4\eps$   & 0            & $1/9+7\eps$ \\
  \hline
  $T_{12}$ & -1& 1& 1& 1& -1& -1\\
  $T_{13}$ & -1& 1& 1& -1& 1& -1\\
  $T_{23}$ & -1& -1& -1& -1& 1& 1\\
  \hline
\end{tabular}
\end{center}
Note that we have
\begin{eqnarray*}
f_1\cdot T_{12} = 7/9-14\eps, & h\cdot T_{12} = -7/9+14\eps, & f_2\cdot T_{12}= -1,\\
f_1\cdot T_{13} = 1/3+30x, & h\cdot T_{13} = -1/3+42x, & f_3\cdot T_{13}= -1 + 36x,\\
f_2\cdot T_{23} = -1/3+60x,& h\cdot T_{23} = -5/9+28x, & f_3\cdot T_{23} = -7/9+14x.
\end{eqnarray*}
Hence $f_1$ wins over $f_3$, $f_3$ wins over $f_2$, and $f_2$ wins over $f_1$. Since $f_3=f'_3$ we
have that $f_1$ is the tournament winner. Finally, we have
$\|f_1-g\|_1=2-72\eps$ and $\|f_2-g\|_1=2/9+32\eps$. As $\eps\rightarrow 0$ the
ratio $\|f_1-g\|_1/\|f_2-g\|_1$ gets arbitrarily close to $9$.
\EPRF

\bibliographystyle{alpha}
\bibliography{devroye}

\end{document}